\documentclass[letterpaper]{article}

\usepackage{natbib,alifeconf}  
\usepackage{amssymb,amsmath,multirow,rotate,color}
\usepackage{url,hyperref,cleveref}
\usepackage{booktabs}

\usepackage{subfig}
\usepackage{float}
\usepackage[export]{adjustbox}
\usepackage{multirow}
\usepackage[table,xcdraw]{xcolor}

\usepackage{array}

\makeatletter
\newcommand{\thickhline}{%
    \noalign {\ifnum 0=`}\fi \hrule height 1pt
    \futurelet \reserved@a \@xhline
}
\newcolumntype{"}{@{\hskip\tabcolsep\vrule width 1pt\hskip\tabcolsep}}
\makeatother

\title{Reservoir Computing with Evolved Critical Neural Cellular Automata}

\author{
    Sidney Pontes-Filho$^{1}$, Stefano Nichele$^{2,3}$ \and Mikkel Lepperød$^{1}$ \\
    \mbox{}\\
    $^1$Department of Numerical Analysis and Scientific Computing, Simula Research Laboratory, Norway \\
    $^2$Department of Computer Science and Communication, Østfold University College, Norway \\
    $^3$Department of Computer Science, Oslo Metropolitan University, Norway \\
    sidneypontesf@gmail.com
} 

\begin{document}

\maketitle

\begin{abstract}
Criticality is a behavioral state in dynamical systems that is known to present the highest computation capabilities, i.e., information transmission, storage, and modification. Therefore, such systems are ideal candidates as a substrate for reservoir computing, a subfield in artificial intelligence. Our choice of a substrate is a cellular automaton (CA) governed by an artificial neural network, also known as neural cellular automaton (NCA). We apply evolution strategy to optimize the NCA to achieve criticality, demonstrated by power law distributions in structures called avalanches. With an evolved critical NCA, the substrate is tested for reservoir computing. Our evaluation of the substrate is performed with two benchmarks, 5-bit memory task and image classification of handwritten digits. The result of the 5-bit memory task achieved a perfect score and the system managed to remember all 5 bits. The result for the image classification task matched and sometimes surpassed the performance of the best elementary CA for this task. Moreover, the proposed critical NCA may operate as a self-organized critical system, due to its robustness to extreme initial conditions.
\end{abstract}


Data/Code available at: \url{https://github.com/bioAI-Oslo/critical-nca-reservoir}

\section{Introduction}

\begin{figure}[t!]
\centering
\includegraphics[width=0.47\textwidth]{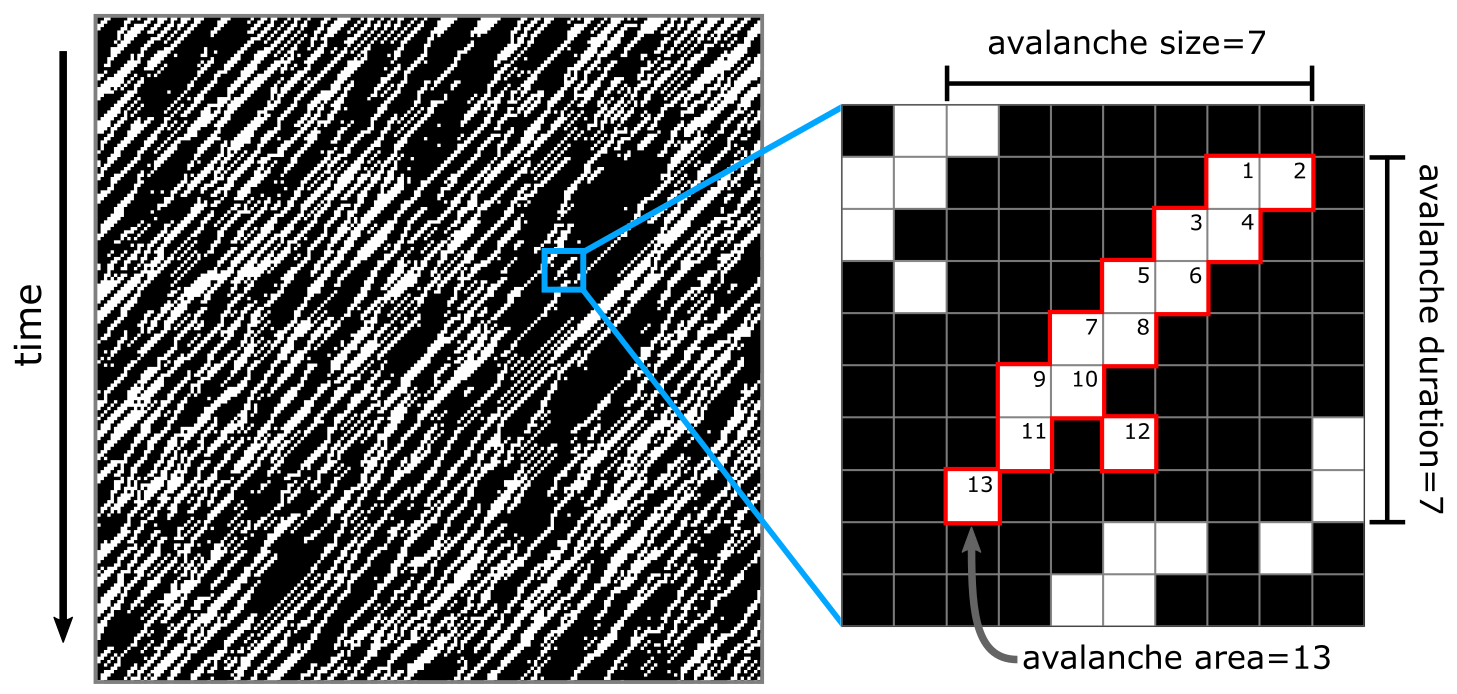}
\caption{Sample of the evolved critical neural cellular automata with 200 cells (horizontal axis) randomly initialized and ran through 200 time-steps (vertical axis from top to bottom). White cells are state 0, and black cells are state 1. The zoomed-in portion to the right includes the definition of size, duration, and area of an avalanche of state 0, which is highlighted by a red border. An avalanche is a cluster of neighboring cells with the same state. The size is identified by the amount of cells affected in an avalanche. The duration informs the quantity of time-steps an avalanche lasted. The area is regarding the size of the cluster in space and time.}
\label{fig:ca}
\end{figure}

Distributed dynamical systems may possess remarkable computational capabilities as a result of their interacting components giving rise to emergent behavior. However, such behavior must possess certain characteristics. One of such key characteristics for computation is criticality, which is known to optimize information storage, transmission, and modification \citep{langton1990computation}. One way to achieve criticality in a dynamical system is by changing its dynamical phase with one or more controlling parameters, such as temperature and pressure. For a system to present critical behavior, the parameters need to be tuned to a point near a phase transition. In this way, the system would have a combination of the two phases in the vicinity of a transition; one being more ordered, and the other more chaotic. Therefore, this state is also referred to as ``at the edge of chaos'', and presents complex structures in space and time. The more ordered phase is responsible for the storage of information because it keeps the information longer within the system, and is sometimes permanent when the system is static; the chaotic phase supports modification of information by frequently changing information in the system. Furthermore, such critical behavior may display the same activity patterns spread in space and time across different scales, similar to a fractal in the space dimension. Thus, such systems have a power law distribution of specific dynamical structures. A power law of the value $x$ is defined by a probability distribution
\begin{equation}
\label{eq:powerlaw}
P(x)\propto x^{-\alpha},
\end{equation}
where $\alpha$ is the exponent or slope of the distribution with common values in a range $(2,3)$, but with exceptions~\citep{clauset2009power}. A usual structure for analyzing criticality is an avalanche, which is identified by a global or local cluster of components in the system with the same state. Most critical systems exhibit these two types of behavior. However, a critical system with a control parameter is rarely observed and can remain in criticality independently of initialization or tuning. The reason behind this is that the critical state is an attractor to the system, and this is known as self-organized criticality (SOC)~\citep{bak1987self,bak1988self,pontes2022assessing}. Natural events of this criticality were detected in some distributed dynamical systems, such as the stock market~\citep{mandelbrodtbook,joao2012,joao2013}, and notably in neuronal avalanches in the cortex~\citep{beggs2003neuronal}. Moreover, SOC is hypothesized to support intelligence in the human brain~\citep{fontenele2019criticality,heiney2021criticality}. This is often referred to as the critical brain hypothesis~\citep{chialvo2004critical,hesse2014self}.

To explore and understand the effects of fractal-like critical behavior in the computational capability of a simple distributed dynamical system, our goal is to optimize a deterministic one-dimensional cellular automaton (CA) toward criticality by finding a transition rule that gives power law distributions for the size, duration, and area of its avalanches (see Fig.~\ref{fig:ca}). A CA system consists of discrete computing units or cells regularly distributed in a grid, commonly with one or two dimensions. Those cells often have binary states that change in discrete time following a transition rule~\citep{wolframbook}. This system can be optimized toward criticality by searching for proper transition rules. After such optimization, the critical CA can be utilized as a substrate for reservoir computing (RC)~\citep{schrauwen2007overview}, a subfield of artificial intelligence. This paradigm utilizes a dynamical system as a reservoir to perform computation by mapping inputs into a higher dimensional space; then, a linear machine learning model is trained to interpret the states of the substrate perturbed by the input data. This combination of RC and CA is known as Reservoir Computing with Cellular Automata (ReCA). One of the main advantages of ReCA is its energy efficiency and, therefore, its significance for Edge AI~\citep{singh2023edge,glover2024reservoir}. In ~\citep{pontes2020neuro}, a 1D CA was optimized with a genetic algorithm towards criticality. However, we verified that such an CA was not effective in RC due to its stochasticity. Thus, the present work aims to optimize a deterministic CA using a similar approach as in~\citep{pontes2020neuro} and then assess its performance as for reservoir computing.

For more variety in the search space for transition rules, we have chosen a neural cellular automaton (NCA) where the transition rule is defined by an artificial neural network~\citep{nichele2017neat,mordvintsev2020growing} and the parameters to be optimized are the weights and biases as in traditional neuroevolution~\citep{stanley2019designing}. Their optimization is performed by a method called covariance matrix adaptation evolution strategy (CMA-ES)~\citep{hansen1996adapting}. To facilitate and simplify the usage of the selected dynamical system in RC, and due to the definition of an avalanche, the NCA is binary, deterministic, and one-dimensional.

With a successfully evolved critical NCA, the verification of its effectiveness as a substrate in RC is achieved by two common benchmarks in ReCA. The first one is the 5-bit memory task~\citep{yilmaz2014reservoir,nichele2017deep,glover2023investigating}, and the second one is the classification of a binarized version of the MNIST handwritten digit dataset~\citep{lecun1998gradient,glover2024reservoir}. The linear machine learning method to be used as a readout layer is a support vector machine (SVM) with linear kernel~\citep{cortes1995support}.

This work contains novelties and contributions in the area of RC, NCA, evolutionary computing, and criticality. It demonstrates the possibility of using evolution strategy in NCA for the search of criticality measured by power law distribution of avalanches. Moreover, there is an assessment of such neural cellular automata in reservoir computing. In essence, it reports the results of evolving an NCA toward criticality and applying it as a reservoir.

\section{Related works}

Neural cellular automata were initially used for morphogenesis purposes and typically optimized with genetic algorithms~\citep{nichele2017neat}. Subsequently, the optimization of NCA was replaced by gradient descent in the work by~\citet{mordvintsev2020growing}, resulting in a new and increasing interest in morphogenesis and development with self-organizing distributed systems. Several works with NCA for morphogenesis appeared, such as for 3D artifacts and machines~\citep{sudhakaran2021growing} and voxel-based soft robots~\citep{horibe2021regenerating}. In addition, NCA has been used for other purposes, for example, to control a cart-pole agent~\citep{variengien2021towards}, to define the weights of an artificial neural network~\citep{najarro2022hypernca}, and to classify hand-written digits~\citep{randazzo2020self}.

On the other hand, reservoir computing started with two concurrent and independent works. One developed the echo state networks that use randomly connected artificial neural networks as reservoirs~\citep{jaeger2001echo}, and the other introduced the liquid state machines where the reservoir setup is similar but uses spiking neurons instead~\citep{maass2002real}. Reservoir computing allows for different distributed dynamical systems to be employed as substrates. Some works explored physical systems, such as the surface of a bucket of water, an analog circuit, a nanomagnet ensemble, and a \emph{in-vitro} culture of biological neurons over microelectrode arrays~\citep{jensen2018computation,tanaka2019recent}. Others explored CA as reservoirs. The first investigation was with Conway's Game of Life and many elementary cellular automata~\citep{yilmaz2014reservoir}. They were tested on 5-bit and 20-bit memory tasks and compared with echo state networks. CA presents some benefits related to the number of operations for solving a task and is more hardware-friendly. Other ReCA works explore different strategies to use CA as a substrate. For example, CA substrates arranged in a deep layered architecture~\citep{nichele2017deep}, elementary CA with a rule for two different regions~\citep{nichele2017reservoir}, and exploration of cell history~\citep{margem2019reservoir}.

One work related to ours is~\citep{guichard2024critically}, where the NCA is trained to control a cart-pole agent as in~\citep{variengien2021towards}. However, their NCA architecture was modified and two pre-training methods were introduced. Those novel pre-training methods lead the NCA toward criticality through the maximization of correlation length. Two methods for achieving such maximization are presented, one is implicit and the other is explicit. They are similar to the original pre-training method that trains the NCA to perform an average of the inputs, where input cells and output cells are permanently fixed in the grid. As such, the implicit pre-training is executed as the original one, but the input cells are randomly located. Therefore, the correlation length increases due to the need to maximize the sensitivity and propagation distance of information given as perturbations in the system. The explicit pre-training calculates the correlation length and maximizes it using gradient descent. After pre-training, the NCA is trained with reinforcement learning as the original, or with evolution strategy. If the NCA was kept frozen during training and coupled with a linear readout layer, such framework would resemble a reservoir computer.

\section{Methods}

Our framework consists of (1) the evolution of a neural cellular automaton toward criticality and (2) the application of the resulting evolved NCA as a substrate in reservoir computing. The details are described in the following subsections.

\subsection{Neural cellular automata architecture}

The proposed NCA is similar to an elementary CA. It is binary, deterministic, one-dimensional and the neighborhood size is equal to 3. The modifications from an elementary CA are that the transition rules are defined by an artificial neural network and there are extra binary channels for more diversity in the dynamics of the system and hence a wider search space to be exploited by evolution.

\begin{figure}[t]
\centering
\includegraphics[width=0.47\textwidth]{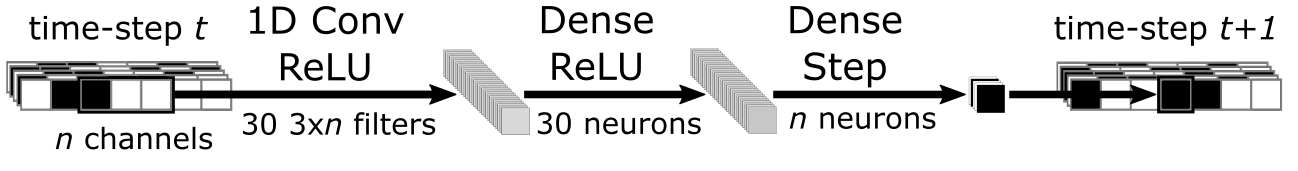}
\caption{Architecture of the one-dimensional neural cellular automata. ``1D Conv'' stands for a 1D convolution layer and Dense is for a fully connected layer. Rectified Linear Unit (ReLU) and Step are activation functions for the layers.}
\label{fig:arch}
\end{figure}

The architecture of the NCA consists of one 1D convolution layer with 30 kernels of size 3 (neighborhood) and rectified linear unit (ReLU) as the activation function~\citep{nair2010rectified}, followed by a dense layer with 30 neurons with ReLU, and a dense layer with step activation function~\citep{sharma2017activation} for the binary states of the cells and the number of neurons equals the amount of channels $n$. When the convolution kernels pass through all possible positions in the CA grid with stride 1 and periodic boundary condition, the CA grid is fully updated. This architecture is illustrated in Fig.~\ref{fig:arch}. In our experiments, we use $n=5$, and the number of cells in the grid depends on the task.

\subsection{Evolution toward Criticality}

The weights and biases of the NCA are evolved with CMA-ES through 100-200 generations with 96 individuals. The fitness function used to guide the evolution is based on the work in~\citep{pontes2020neuro,pontes2022assessing}. An individual NCA is initialized with all binary cells in state $0$ except the first channel, which is randomly initialized. This NCA is simulated through 1,000 time-steps. After that, the first channel is taken and produces a $1,000\times 1,000$ matrix where the distributions of the six avalanche types are acquired, which are size, duration, and area for states 0 and 1. An avalanche is a cluster of cells with the same state. Our definition considers a neighborhood radius for an avalanche of one cell and one time-step. In other words, a $3\times 3$ neighborhood in a space and time visualization. This visualization and the definition of an avalanche are shown in Fig.~\ref{fig:ca}. With those six avalanche distributions, the fitness function calculates the similarities with their power law distributions estimated with a linear fitting method called least squares regression applied to the 10 leftmost bins of the distribution. For that, more than five non-zero bins in the 10 leftmost bins of all distributions are required. Otherwise, the final fitness score $S$ is zero. After a valid power law model estimation, the similarity measurements are the six coefficients of determination of complete linear fitting $\pmb{R^2}$~\citep{wright1921correlation} and the six coefficients of the Kolmogorov-Smirnov (KS) statistic $\pmb{D}$~\citep{clauset2009power}, one measurement for each avalanche distribution and their math symbols are in bold because they are an array, in this case containing six values. Additional fitness measurements are the six percentages of non-zero bins in the distribution $\pmb{B}$, the percentage of unique states of the system during simulation $U$, and the six log-likelihood ratios of the comparison between the power law model and the exponential model for estimating the avalanche distributions $\pmb{L}$. The conversions of the arrays $\pmb{R^2}$, $\pmb{D}$, $\pmb{B}$, and $\pmb{L}$ to single score values are done with the following equations:

\begin{equation}
\label{eq:r2}
\hat{R^2} = \text{sigmoid}(\alpha_{R^2}(0.9\max(\bar{R^{2}_{0}}, \bar{R^{2}_{1}})+0.1\bar{R^{2}})),
\end{equation}

\begin{equation}
\label{eq:d}
\hat{D} = \exp(\alpha_D(0.9\min(\bar{D_{0}}, \bar{D_{1}})+0.1\bar{D})),
\end{equation}

\begin{equation}
\label{eq:b}
\hat{B} = \text{tanh}(\alpha_{B}(0.9\max(\bar{B_{0}}, \bar{B_{1}})+0.1\bar{B})),
\end{equation}

\begin{equation}
\label{eq:l}
\hat{L} = \text{sigmoid}(\alpha_{L}(0.9\max(\bar{L_{0}}, \bar{L_{1}})+0.1\bar{L})).
\end{equation}

In \eqref{eq:r2}, $\hat{R^2}$ is the resulting single value for the fitness score related to the coefficient of determination. $\bar{R^{2}}$ is the average coefficient of determination for all six avalanche distributions, while $\bar{R^{2}_0}$ and $\bar{R^{2}_1}$ are the averages for the three distributions for state 0 and state 1, respectively. $\alpha_{R^2}$ is a smoothing parameter for $R^2$. The same principle applies to \eqref{eq:d} - \eqref{eq:l}. The values of the smoothing parameters are $\alpha_{R^2}=0.01$, $\alpha_{D}=1$, $\alpha_{B}=5$, and $\alpha_{L}=0.01$. The equations to calculate a single value for the fitness scores are meant to map measurements that can go toward negative or positive infinity, so they can be in a certain range, such as [0,1]. That is the reason for using the sigmoid function in \eqref{eq:r2} and \eqref{eq:l}. The use of a hyperbolic tangent in \eqref{eq:b} and an exponential function in \eqref{eq:d} is due to the minimum value being zero. The smoothing parameter $\alpha$ regulates the steepness of those functions for a better fit to the common values of the raw measurements. The maximum value of these scores is 1 and their equations are defined heuristically to give more importance to the avalanche state with a better average measurement (multiplied by $0.9$) relative to their overall average (multiplied by $0.1$). With that, the final fitness score $S$ can be calculated with:

\begin{equation}
\label{eq:first_fit}
S_{partial} = (\hat{R^2})^2 
   + \hat{D}^2  
   + \hat{B} 
   + U,
\end{equation}
\begin{equation}
\label{eq:final_fit}
S = \begin{cases}
S_{partial}+\hat{L}, &S_{partial}>3.0\\
S_{partial}, &\text{otherwise.}
\end{cases}
\end{equation}

The fitness scores $\hat{R^2}$ and $\hat{D}$ are squared in \eqref{eq:first_fit} because they are the scores that measure similarity to a power law distribution, which is the most important measurement for the fitness function. In \eqref{eq:final_fit}, the final score for log-likelihood ratio $\hat{L}$ is only considered when $S_{partial}>3.0$ because of its intensive and slow process for calculating the six ratios. Therefore, the other fitness scores need to be good enough, so the log-likelihood ratios can be measured. A measurement of the six ratios for $\pmb{L}$ is only valid if it presents a $p$-value lower than $0.1$, and therefore considered trustworthy. Otherwise, the measurement of such a ratio is set to $0$.

To verify the presence of criticality in an evolved NCA, the selected individual passes through the goodness-of-fit test based on the Kolmogorov-Smirnov statistic of the six avalanches~\citep{clauset2009power,pontes2020neuro}. This test calculates a $p$-value by comparing the KS statistic of the original distribution with other 1,000 distributions with 10,000 samples. Those distributions are randomly generated using a power law model estimated by the maximum likelihood estimation method with the minimum value on the horizontal axis being set to 1. The $p$-value is the percentage of the KS statistic of the original distribution being better than the statistics of the other generated distributions. \citet{clauset2009power} consider a distribution to be a power law when $p$-value is greater than $0.1$. Calculations of the log-likelihood ratios in $\pmb{L}$, and the goodness-of-fit tests are performed using the \emph{powerlaw} Python library~\citep{alstott2014powerlaw}.

\subsection{Reservoir computing with NCA}

After evolution, an optimized NCA toward criticality that passes the goodness-of-fit test is used as a substrate for reservoir computing. This is to assess the effectiveness of a critical NCA in RC. There are two RC benchmarks where the substrate is evaluated. One benchmark is the 5-bit memory task. It verifies the capability of a reservoir to retain information and remember it after a distractor period. This information is 5 bits that are given sequentially in every input step of the reservoir. The distractor period in our experiment is 200 time-steps for the reservoir. For every time-step of the reservoir, there are three NCA time-steps, and the input is received in the first time-step and in the first channel through a XOR logical operation. The final distractor time-step is a cue signal, so the next 5 time-steps are used to read the reservoir state and remember the 5 input bits in sequence given in the beginning of the task. An example of the task is presented in Table~\ref{tab:memory}. The 5-bit input to be memorized is $10110_{2}$ or $22_{10}$. The reservoir receives four binary inputs. They are the original input, its logical negation, a signal that all input bits have been given, and a cue signal to start the recall of the input bits. The linear readout layer for the reservoir has three outputs, which are the prediction of the original input, its logical negation, and a waiting signal. In our experiment, the grid of the NCA is split into four regions of size 20. Hence, the total grid size is 80. Each region receives the four inputs in random locations. We run this experiment 100 times with different random locations every time, and test it with all 32 combinations of 5 binary numbers. A similar setup was used in the work by~\citep{glover2023investigating} and their GitHub repository was used as basis for our 5-bit memory task implementation.

\begin{table}[!ht]
\begin{tabular}{|l|llll|lll|l|}
\hline \thickhline
\multicolumn{1}{|c|}{Step} & \multicolumn{4}{c|}{Input}                                    & \multicolumn{3}{c|}{Output}                               & \multicolumn{1}{c|}{Stage}          \\ \hline \thickhline
1                          & \cellcolor[HTML]{D1D1D1}1 & \cellcolor[HTML]{D1D1D1}0 & 0 & 0 & 0                         & 0                         & 1 &                                     \\ \cline{1-8}
2                          & \cellcolor[HTML]{D1D1D1}0 & \cellcolor[HTML]{D1D1D1}1 & 0 & 0 & 0                         & 0                         & 1 &                                     \\ \cline{1-8}
3                          & \cellcolor[HTML]{D1D1D1}1 & \cellcolor[HTML]{D1D1D1}0 & 0 & 0 & 0                         & 0                         & 1 &                                     \\ \cline{1-8}
4                          & \cellcolor[HTML]{D1D1D1}1 & \cellcolor[HTML]{D1D1D1}0 & 0 & 0 & 0                         & 0                         & 1 &                                     \\ \cline{1-8}
5                          & \cellcolor[HTML]{D1D1D1}0 & \cellcolor[HTML]{D1D1D1}1 & 0 & 0 & 0                         & 0                         & 1 & \multirow{-5}{*}{Bits to memorize}  \\ \hline \thickhline
6                          & 0                         & 0                         & 1 & 0 & 0                         & 0                         & 1 &                                     \\ \cline{1-8}
...                        & 0                         & 0                         & 1 & 0 & 0                         & 0                         & 1 &                                     \\ \cline{1-8}
204                        & 0                         & 0                         & 1 & 0 & 0                         & 0                         & 1 & \multirow{-3}{*}{Distractor period} \\ \hline \thickhline
205                        & 0                         & 0                         & 0 & 1 & 0                         & 0                         & 1 & Cue signal                          \\ \hline \thickhline
206                        & 0                         & 0                         & 1 & 0 & \cellcolor[HTML]{D1D1D1}1 & \cellcolor[HTML]{D1D1D1}0 & 0 &                                     \\ \cline{1-8}
207                        & 0                         & 0                         & 1 & 0 & \cellcolor[HTML]{D1D1D1}0 & \cellcolor[HTML]{D1D1D1}1 & 0 &                                     \\ \cline{1-8}
208                        & 0                         & 0                         & 1 & 0 & \cellcolor[HTML]{D1D1D1}1 & \cellcolor[HTML]{D1D1D1}0 & 0 &                                     \\ \cline{1-8}
209                        & 0                         & 0                         & 1 & 0 & \cellcolor[HTML]{D1D1D1}1 & \cellcolor[HTML]{D1D1D1}0 & 0 &                                     \\ \cline{1-8}
210                        & 0                         & 0                         & 1 & 0 & \cellcolor[HTML]{D1D1D1}0 & \cellcolor[HTML]{D1D1D1}1 & 0 & \multirow{-5}{*}{Bits to recall}    \\ \hline \thickhline
\end{tabular}
\caption{Example of the 5-bit memory task to memorize the number 22 in binary after a distractor period of 200 time-steps. The gray background indicates the bits to be memorized and recalled later. Adapted from~\citep{babson2019reservoir,glover2023investigating}.}
\label{tab:memory}
\end{table}

The other benchmark is the image classification of the MNIST handwritten digit dataset. It consists of grayscale images with size $28\times 28$ with the 10 digits, from 0 to 9, as classes. Since the reservoir is binary, the images are binarized with a pixel intensity threshold of $0.5$ when the range of pixel intensity is $[0,1]$. The training set contains 60,000 examples and the test set 10,000 examples, as the original dataset~\citep{lecun1998gradient}. For reservoir computing, the image is flattened into an array of size 784, which is also the number of cells in the NCA used as a substrate. The first channel of the NCA is initialized with the flattened image, four time-steps are run, and the readout layer classifies the input image using the states of all channels in all time-steps. There are five channels, four time-steps, and 784 cells; therefore, the readout layer receives 15,680 binary values. The configuration of four CA time-steps used for classification is the same as in~\citep{glover2024reservoir} in order to be able to compare their results with ours. 

The readout layer for the two tasks is implemented with a support vector machine with linear kernel~\citep{cortes1995support}. The hyperparameters of the linear SVM are the default of the \emph{scikit-learn} Python library~\citep{pedregosa2011scikit}. 

\section{Results}

\begin{figure}[ht]
\centering
\includegraphics[width=0.47\textwidth]{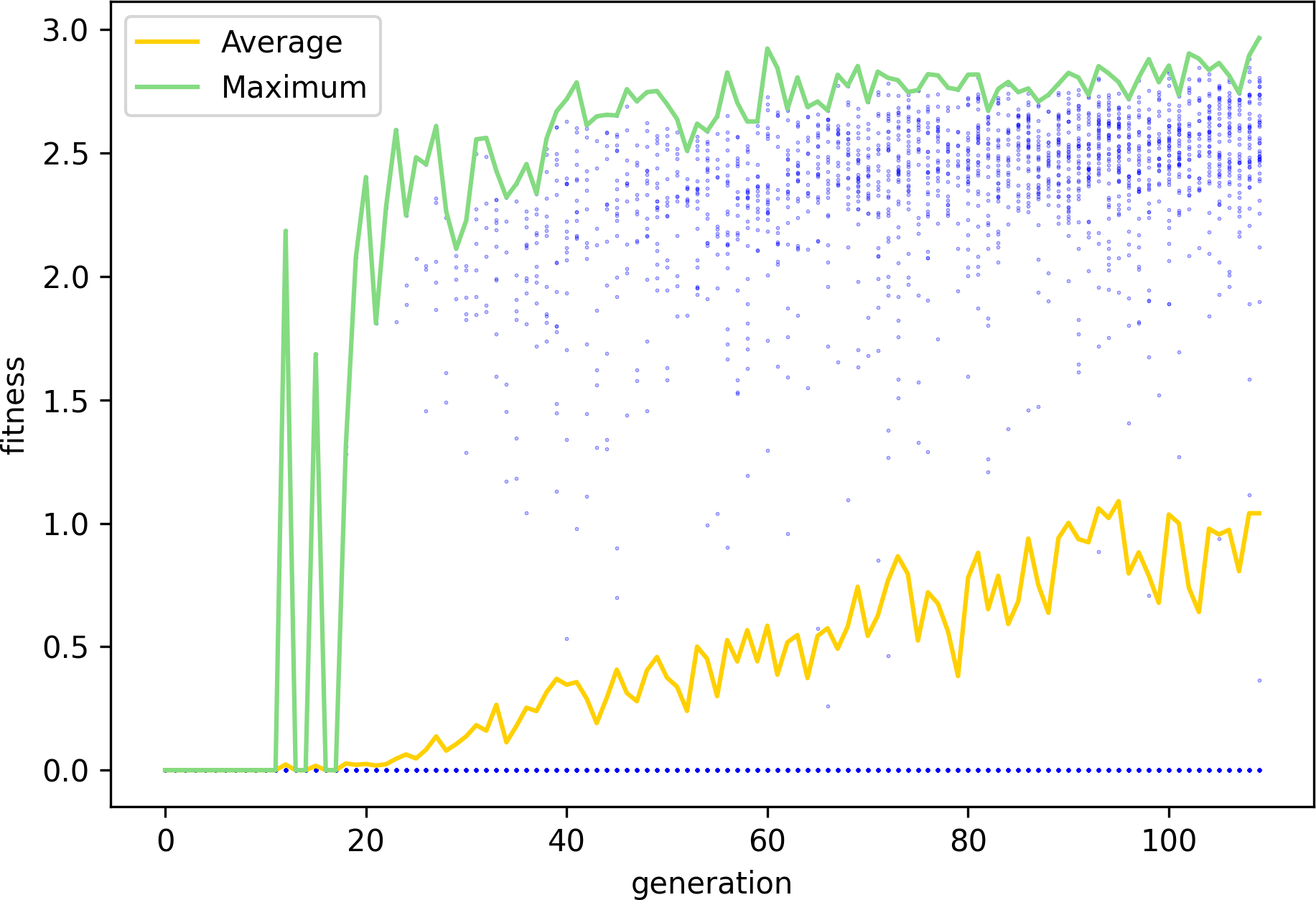}
\caption{Fitness history of the evolution that generated the selected evolved NCA in generation 110, the last one. The blue dots are the final fitness scores of the individuals.}
\label{fig:fitness}
\end{figure}

\begin{figure*}[!ht]
\centering
\subfloat[Avalanche size of state 0]{\label{fig:distribution1}\includegraphics[width=0.33\textwidth]{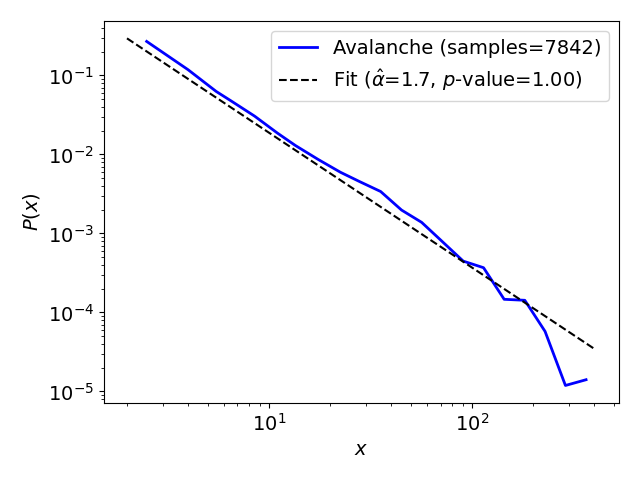}}\hfill
\subfloat[Avalanche duration of state 0]{\label{fig:distribution2}\includegraphics[width=0.33\textwidth]{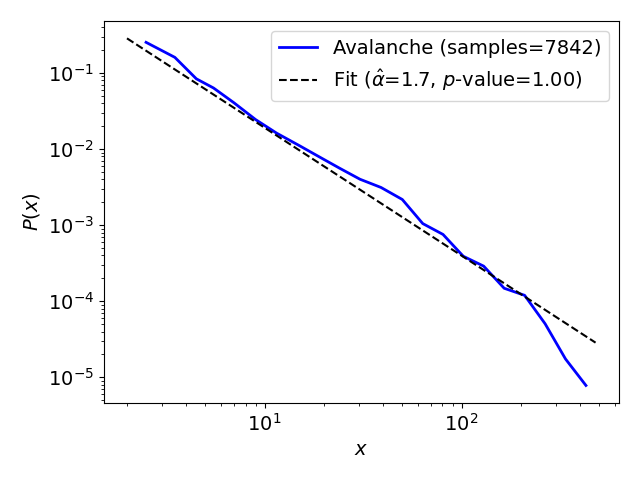}}\hfill
\subfloat[Avalanche area of state 0]{\label{fig:distribution3}\includegraphics[width=0.33\textwidth]{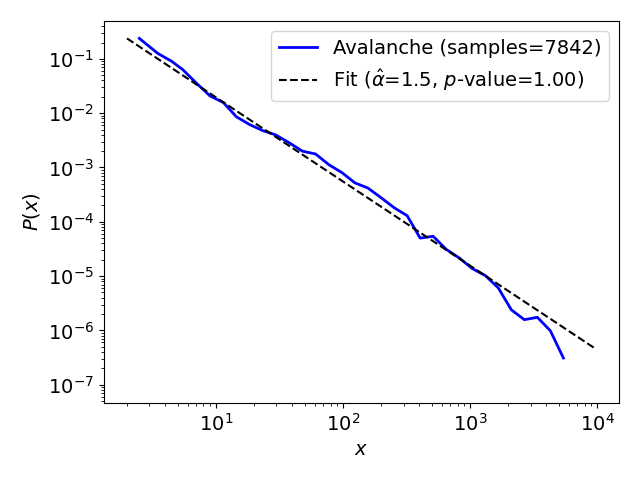}}\\
\subfloat[Avalanche size of state 1]{\label{fig:distribution4}\includegraphics[width=0.33\textwidth]{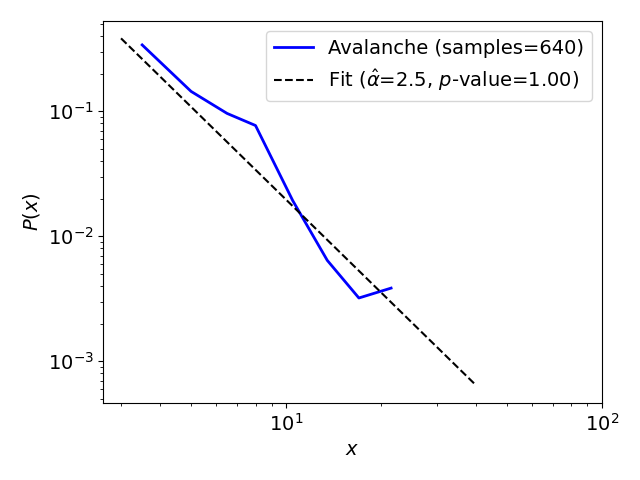}}\hfill
\subfloat[Avalanche duration of state 1]{\label{fig:distribution5}\includegraphics[width=0.33\textwidth]{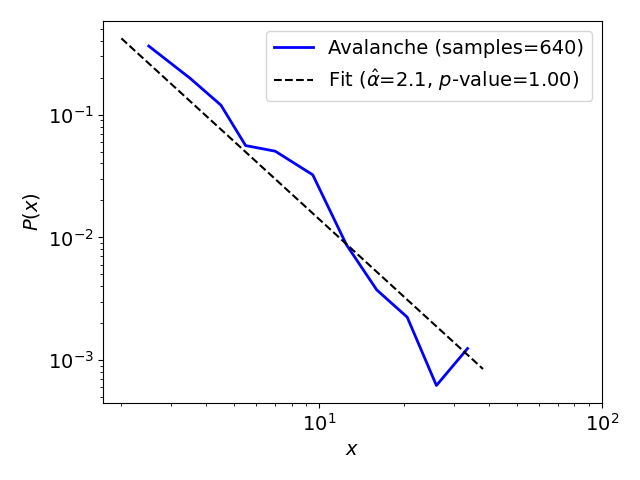}}\hfill
\subfloat[Avalanche area of state 1]{\label{fig:distribution6}\includegraphics[width=0.33\textwidth]{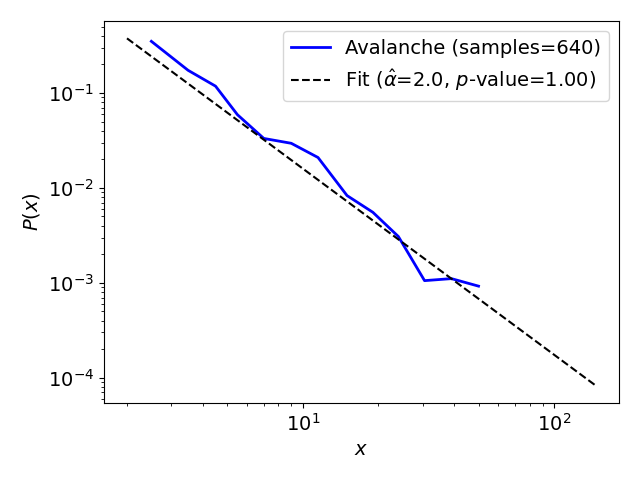}}
\caption{Example of the avalanche distributions of the selected NCA. The estimated power law slope $\hat{\alpha}$ and goodness-of-fit $p$-value are included in the legends.}
\label{fig:distribution}
\end{figure*}

The evolution toward criticality in an NCA required to be executed a few times until the goodness-of-fit $p$ value of the six avalanche distributions was greater than $0.1$. This run had 110 generations and its fitness history is depicted in Fig.~\ref{fig:fitness}. The fittest individual is shown in Fig.~\ref{fig:ca} and has a final fitness score $S$ of 2.9656. The scores used to calculate it are shown in Table~\ref{tab:fitness}. Since the fittest individual did not pass the threshold of 3.0 in the partial fitness score $S_{partial}$, the log-likelihood ratio $\hat{L}$ that compares the exponential model with the power law one was not required in this case. However, the six avalanche distributions got a goodness-of-fit $p$-value equal to 1.0, and qualitatively similar to an estimated power law model by the maximum likelihood estimation method. This can be seen in Fig.~\ref{fig:distribution} together with their estimated power law slope $\hat{\alpha}$ and goodness-of-fit $p$-value. Therefore, this evolved NCA is considered to possess a critical behavior.

\begin{table}[hb]
\centering
\begin{tabular}{ll}
\toprule
Fitness score                               & Value  \\
\midrule
Coefficients of determination $\hat{R^2}$   & 0.6909 \\
Kolmogorov-Smirnov statistic $\hat{D}$ & 0.9632 \\
Percentages of non-zero bins $\hat{B}$      & 0.5603 \\
Percentage of unique states $U$             & 1.0    \\
Log-likelihood ratio $\hat{L}$              & -      \\ \midrule
Final fitness score $S$                     & 2.9656 \\
\bottomrule
\end{tabular}
\caption{Fitness scores of the selected NCA. Extra decimals of the values are hidden for avoiding clutter.}
\label{tab:fitness}
\end{table}

\begin{figure*}[!ht]
\centering
\subfloat[100\% state 0 -- 0\% state 1]{\label{fig:nca_test1}\includegraphics[width=0.23\textwidth,frame]{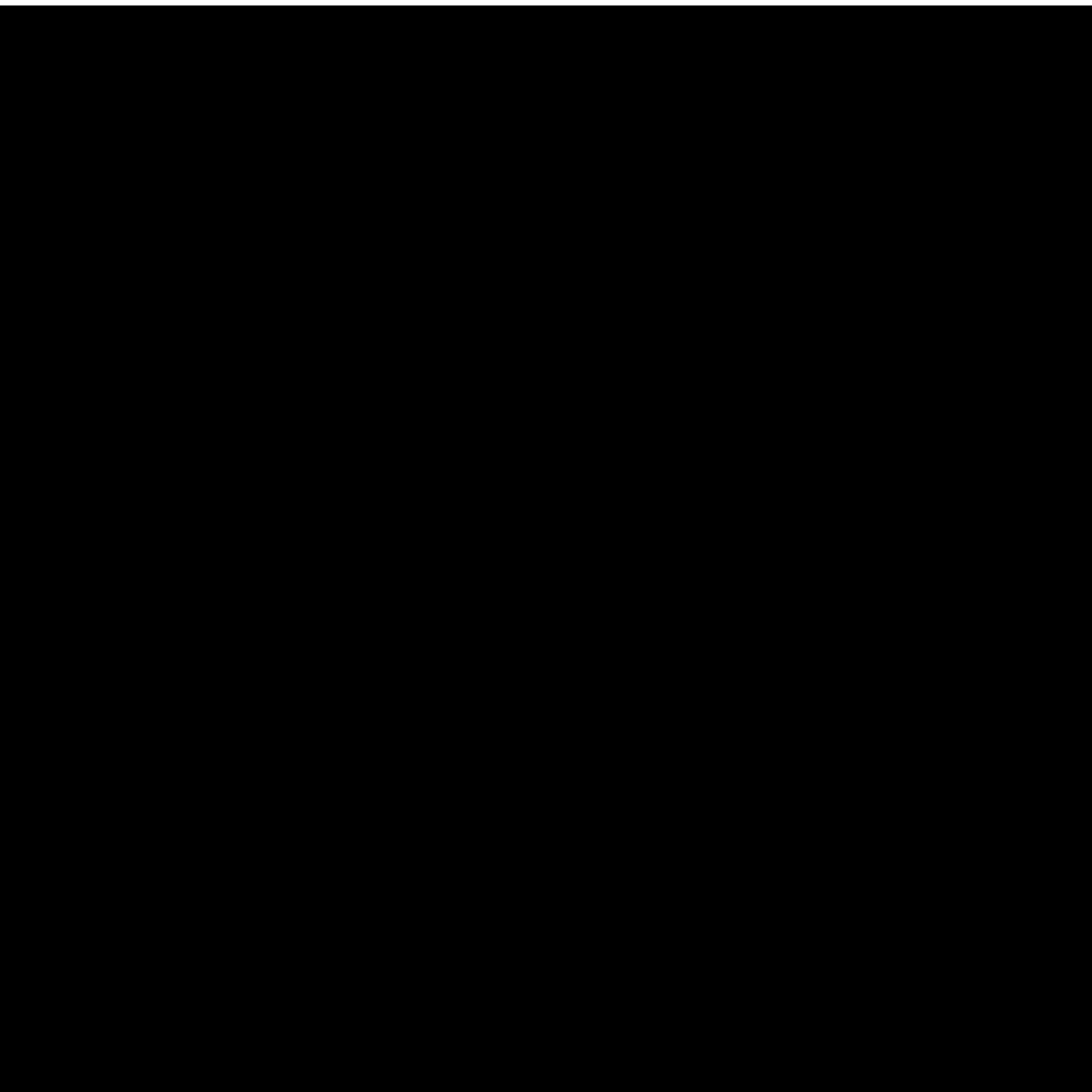}}\hfill
\subfloat[99\% state 0 -- 1\% state 1]{\label{fig:nca_test2}\includegraphics[width=0.23\textwidth,frame]{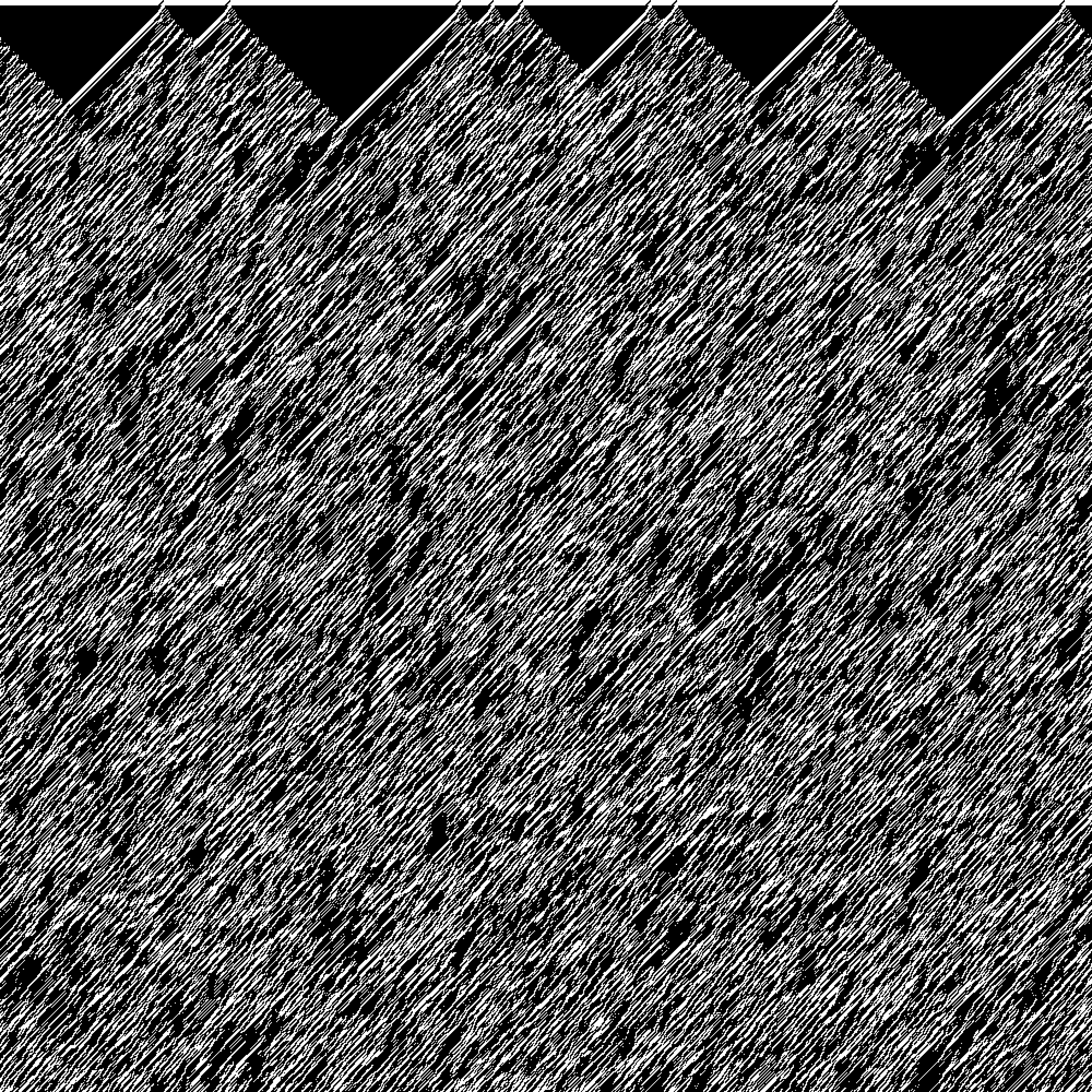}}\hfill
\subfloat[1\% state 0 -- 99\% state 1]{\label{fig:nca_test3}\includegraphics[width=0.23\textwidth,frame]{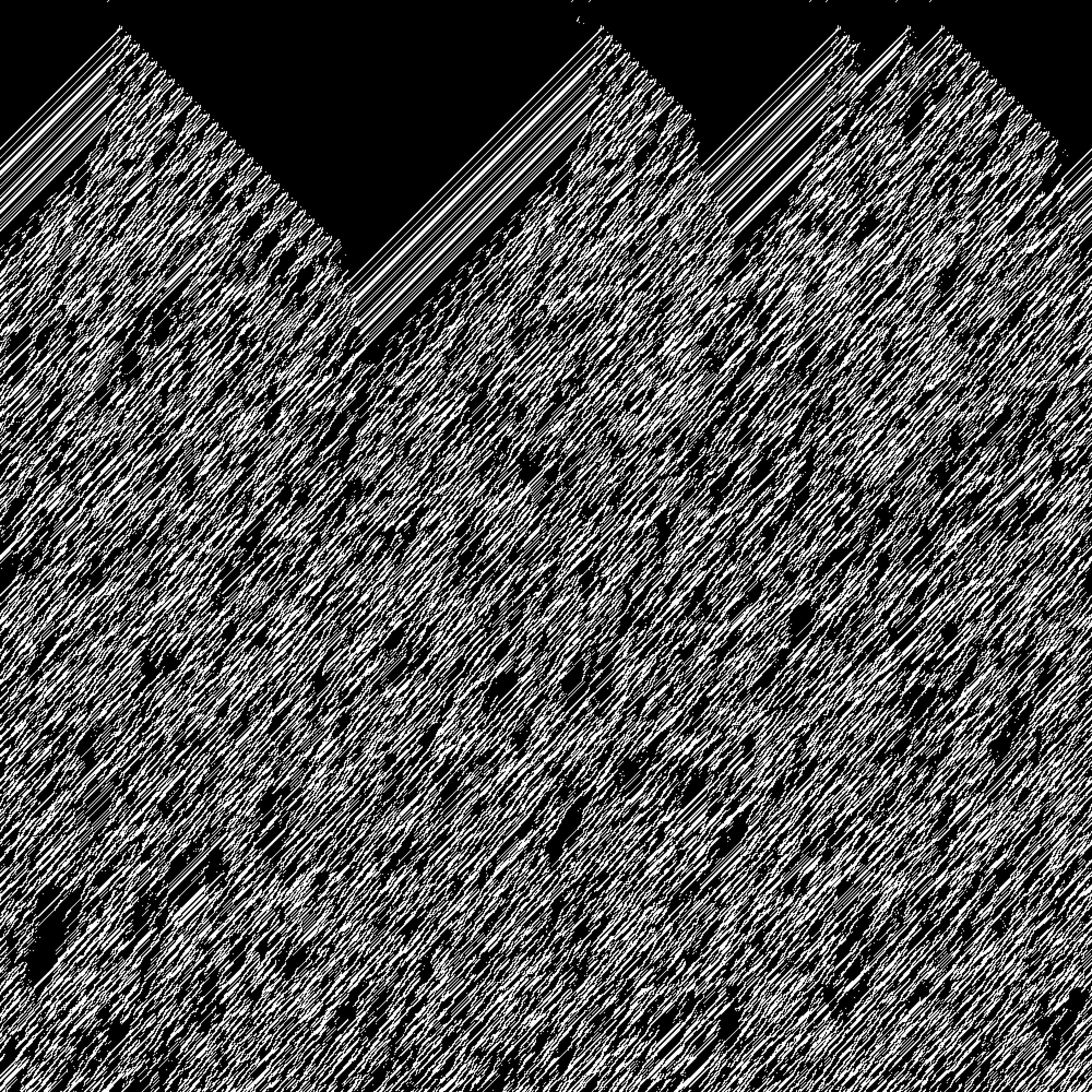}}\hfill
\subfloat[0\% state 0 -- 100\% state 1]{\label{fig:nca_test4}\includegraphics[width=0.23\textwidth,frame]{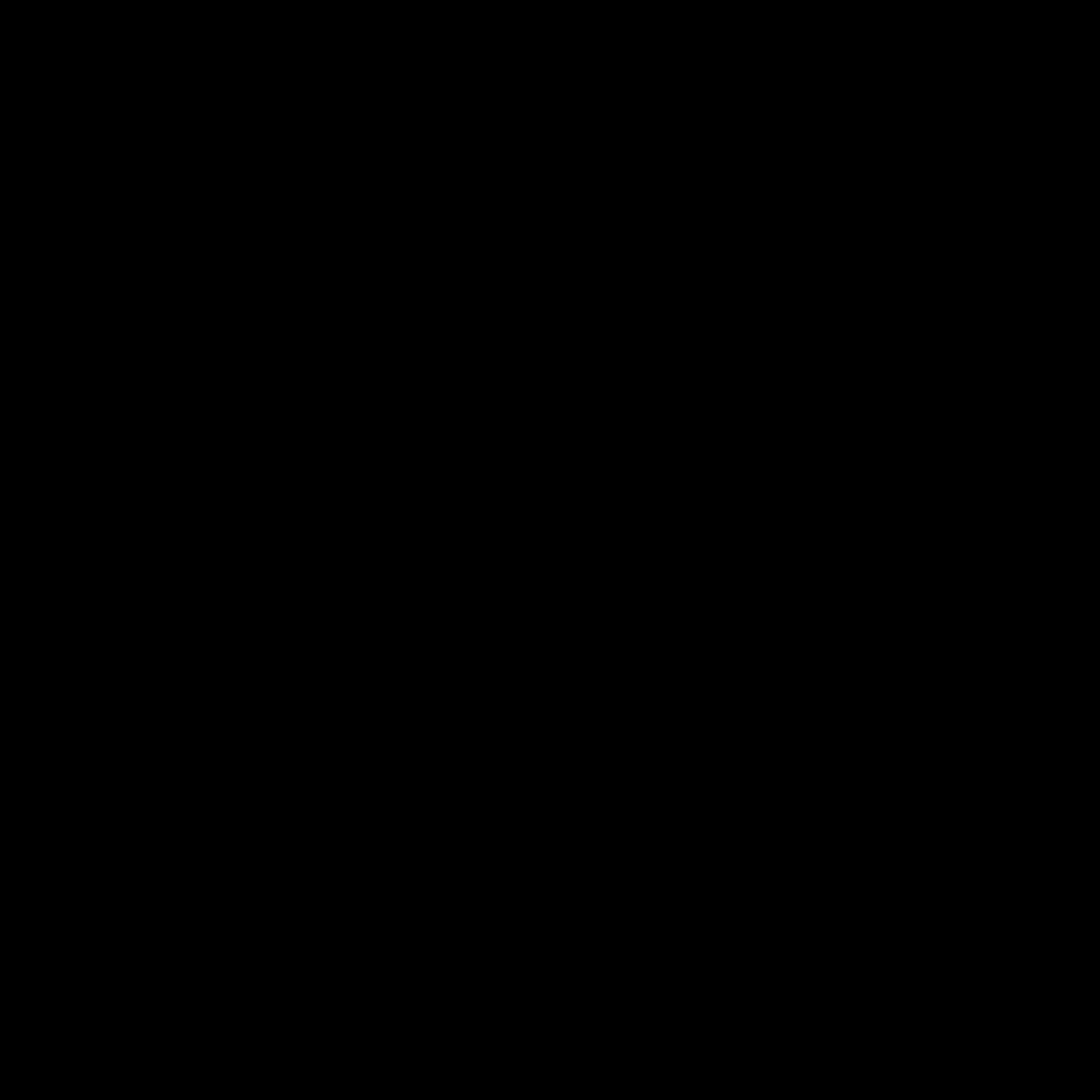}}
\caption{Analysis of cell state initialization in an NCA with 1,000 cells during 1,000 time-steps. Only the first channel is shown.}
\label{fig:nca_test}
\end{figure*}

To test the robustness of the selected critical NCA regarding the uniform initialization during evolution, we evaluate this NCA with extreme initial conditions, such as all cells with state 0, all cells with state 1, 99\% probability of the cell state being 0, and 99\% probability of the cell state being 0. The simulation has the same properties as in the evolution process. So, the NCA has 1,000 cells simulated through 1,000 time-steps. Fig.~\ref{fig:nca_test} shows the effects of these four extreme initial conditions. When the NCA is initialized with only one state, as in Fig.~\ref{fig:nca_test1} and Fig.~\ref{fig:nca_test4}, the state 1 takes over all the following time-steps. When the first time-step contains around 1\% of the cells in state 1, the NCA recovers after a few time-steps its behavior when the initialization is uniform as in Fig.~\ref{fig:ca}. The same happens when the cells start with 1\% probability of state 0 as shown in Fig.~\ref{fig:nca_test3}. In this case, the state 0 (white pixels) disappears and appears again after a short period. However, there are other channels that have maintained the information of the small number of cells initialized with state 0. Then recovering the critical behavior, as it was initialized with equal chances. This recovery of the NCA toward the same behavior that generated all the power law distributions shown in Fig.~\ref{fig:distribution} can be seen as self-organized criticality. This constitutes an attractor toward a critical point independent of the initial conditions of the system. Although this result seems promising, further analysis of this claim is required.

\begin{figure*}[!ht]
\centering
\subfloat[Channel \#1]{\label{fig:memory1}\includegraphics[width=0.18\textwidth,frame]{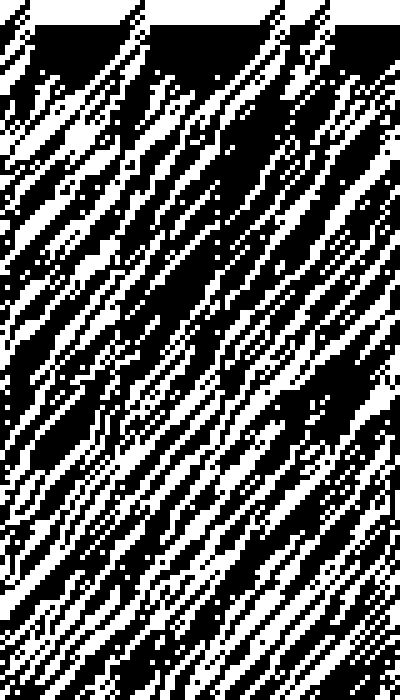}}\hfill
\subfloat[Channel \#2]{\label{fig:memory2}\includegraphics[width=0.18\textwidth,frame]{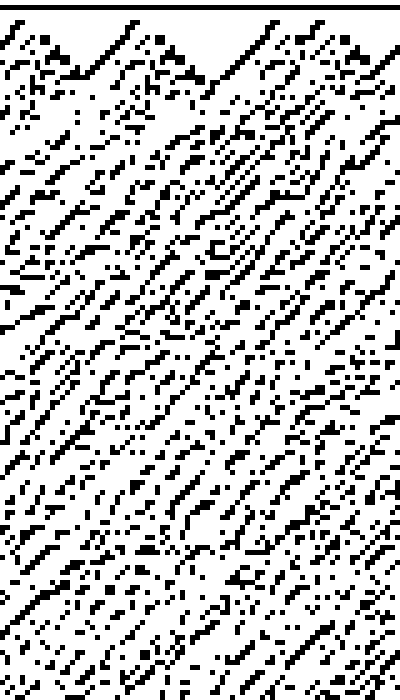}}\hfill
\subfloat[Channel \#3]{\label{fig:memory3}\includegraphics[width=0.18\textwidth,frame]{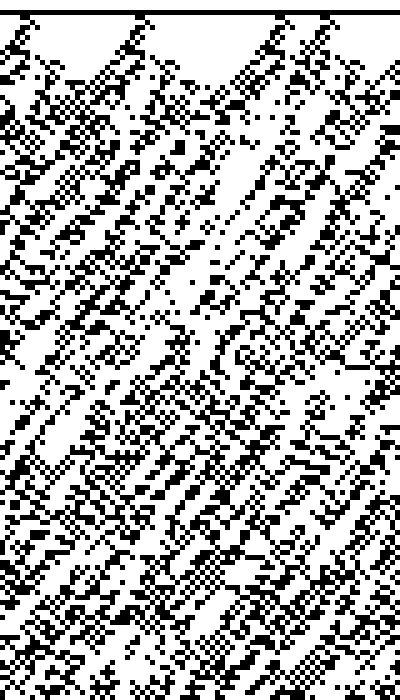}}\hfill
\subfloat[Channel \#4]{\label{fig:memory4}\includegraphics[width=0.18\textwidth,frame]{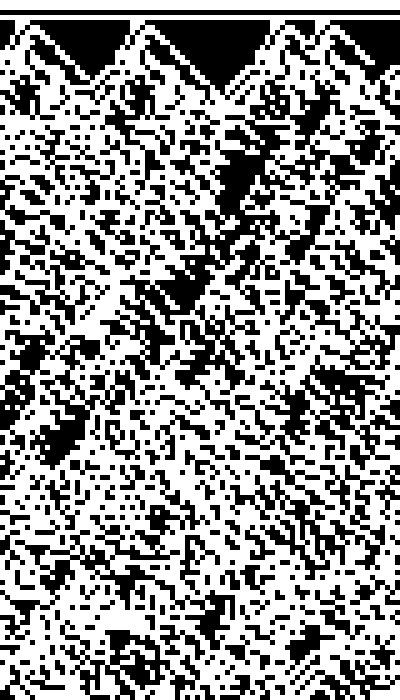}}\hfill
\subfloat[Channel \#5]{\label{fig:memory5}\includegraphics[width=0.18\textwidth,frame]{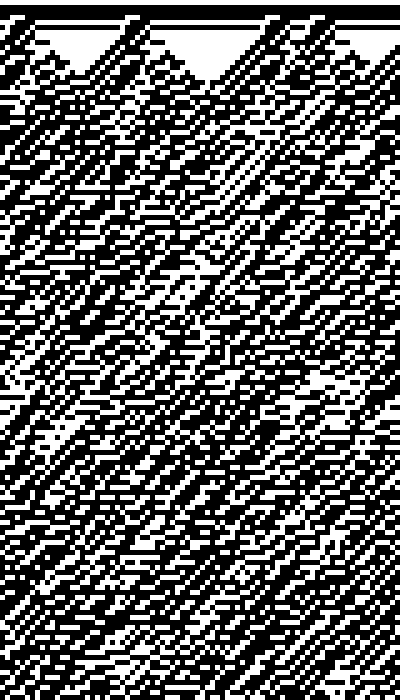}}
\caption{Example of five channels of the evolved NCA executing the 5-bit memory task. This is the first 140 CA time-steps out of 630 (three NCA time-steps per reservoir time-step).}
\label{fig:memory}
\end{figure*}

The application of the selected NCA for reservoir computing using the 5-bit memory benchmark resulted in perfect performance. All 100 runs with different input locations each run recalled all 5 bits and recognized all stages in every time-step correctly. An example of a run of the NCA for the 5-bit memory task is shown in Fig.~\ref{fig:memory}. The inputs are perceivable as being added every three time-steps to the first channel as seen in Fig.~\ref{fig:memory1} with the vertical dotted lines of white color (state 0).

\begin{figure}[t]
\centering
\includegraphics[width=0.47\textwidth]{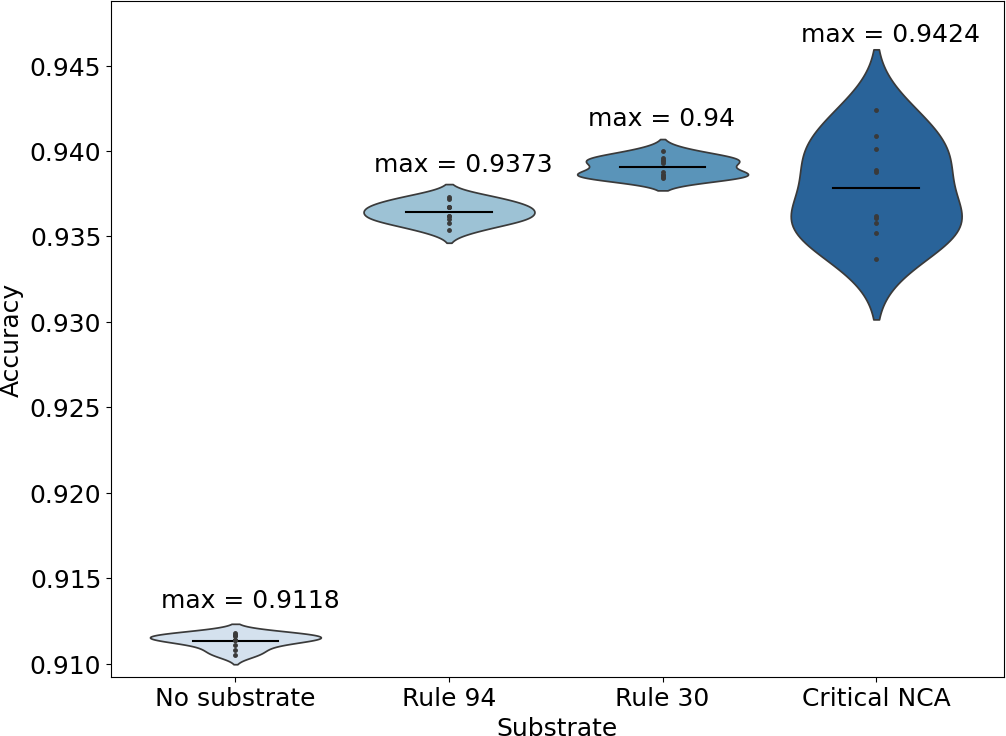}
\caption{Accuracy on the test set of MNIST dataset. Each substrate type was executed 10 times. The black dots indicate those executions. The maximum value of each substrate is marked on top of their accuracies and distributions. The average is represented by the horizontal line.}
\label{fig:mnist}
\end{figure}

The performance results of the critical NCA applied in the image classification task with the MNIST dataset are shown in Fig.~\ref{fig:mnist} as test set accuracy measurements of 10 runs. They are compared with the accuracies of a classifier without substrate (the linear SVM alone classified the binarized images), and the best two rules for an elementary CA applied in ReCA reported by~\citet{glover2024reservoir}, which are rules 94 and 30. The maximum accuracy obtained by the linear readout layer without substrate is 0.9118. With the usage of a substrate, the maximum accuracy increases to 0.9373 of the elementary CA with rule 94 as a substrate, then 0.94 of rule 30, and 0.9424 of our evolved critical NCA. The critical NCA has the highest accuracy, but it presents the highest standard deviation, therefore instability. The average accuracy of the NCA was also lower compared to the elementary CA with rule 30. The NCA has an average of 0.9378, while rule 30 achieved 0.9391. Even though the average accuracy of the critical NCA is lower than the one of the elementary CA with rule 30, the NCA has 3 out of 10 runs with an accuracy higher than the maximum of rule 30.

\section{Discussion}

The results of the evolution toward criticality were successful. All six avalanche distributions presented goodness-of-fit $p$-values equal to 1.0, thus indicating they are all power laws. Additionally, we were able to generate a critical NCA that presents a behavior peculiar to self-organized criticality. Such NCA has a property that other self-organized critical systems do not necessarily show, that is, criticality is self-sustained. It requires a single disturbance to the system for running indefinitely without reaching a quiescent state, unlike other CA models, such as the sandpile model~\citep{bak1987self,bak1988self}, Conway's Game of Life~\citep{bak1989self}, and the evolved stochastic CA in~\citep{pontes2020neuro,pontes2022assessing} that used similar fitness function. Such works require constant or regular perturbations in the system to maintain their activity.

In the MNIST classification, the high standard deviation of the performance of the critical NCA as a substrate may be due to its high dimensionality, since it requires additional channels. The linear SVM receives 15,680 binary values from the critical NCA, while the elementary CA presents 3,136, which is 5 times less.

Efforts were made to try to use only the first channel as a substrate in ReCA, which is the channel evolved to be critical. However, including all 5 channels together presented significant computational benefits in the two benchmarks. All channels presented a power law distribution in at least one state, except channel \#2 depicted in Fig.~\ref{fig:memory2}. An evaluation of the combinations of the channels is required to identify the channels responsible for the computational capabilities of this substrate in ReCA. This paper, due to space constraints, does not present such an analysis.

Some beneficial features of ReCA are energy efficiency and hardware implementation for edge devices~\citep{kalapothas2022efficient,singh2023edge,glover2023investigating}, but the usage of an artificial neural network as a transition rule in a CA may reduce these benefits. Although, the transition rule produced by a binary and deterministic NCA can be converted to a traditional lookup table; in our case with a binary one-dimensional NCA with 3 neighbors and 5 channels, the size of the table, which maps $\{0,1\}^{3\times 5}\to \{0,1\}^5$, would be $2^{3\times 5}=32,768$. Therefore, the presented critical NCA may be implemented in hardware as the physical NCA substrate for 2D shape classification presented in the work by~\citet{walker2022physical}.

\section{Conclusion}

This work presents a framework that applies evolution strategy for optimizing a deterministic, one-dimensional, and binary NCA towards criticality. As such, the resulting NCA can be utilized as a substrate in reservoir computing, as typically done in ReCA. The test of the evolved and selected NCA in the 5-bit memory task was successful and resulted in perfect scores in all 100 runs with different input locations in every run. The application to classify the MNIST dataset presents that the critical NCA achieved test set accuracy measurements similar to the best elementary CA rules and surpasses in some runs the performance of the best elementary CA, i.e., rule 30.

We envisage that NCA can be leveraged as a suitable substrate for reservoir computing and that future works will continue on this path. Our ideas for further development are the substitution of the step function to a sigmoid function of the critical NCA, and benchmark its performance. Additionally, it would be interesting to apply the optimization method for criticality presented in~\citep{guichard2024critically} and evaluate it as a substrate for reservoir computing. Finally, a thorough analysis of the critical NCA as a possible system with self-organized criticality would further strengthen our results.

\section{Acknowledgements}

This paper has benefited from the Experimental Infrastructure for Exploration of Exascale Computing (eX3), which is a computational cluster funded by the Research Council of Norway under contract 270053.

\footnotesize
\bibliographystyle{apalike}
\bibliography{refs} 

\end{document}